# Flow-by-Flow: Content-Judgment Bypass for Governing AI Output in High-Loss Domains

**Hiroki Naito**[1]
[1] **UTIE Research Institute, UTIE Instruments Inc., Japan**

## Abstract

Human review is a load-bearing safeguard in AI governance: oversight mandates such as Article 14 of the EU AI Act presume that accountable reviewers can scrutinize what they approve. In high-loss domains open to any submitter, such as patent examination, litigation and peer review, generative AI allows the number of submissions and the review effort each requires to grow faster than review capacity. This paper asks how institutions can keep this supervisory load within capacity without judging content, since content judgment either reintroduces hallucination risk or consumes scarce review capacity. Treating supervisory load as submission volume times per-item load, we show that if it grows at a compound rate while capacity grows linearly, capacity is exceeded in finite time; added capacity or one-off load reductions postpone exceedance only logarithmically, whereas lowering the growth rate postpones it hyperbolically. We then propose Flow-by-Flow, which prices supervisory load from countable features of submissions and caps the load admitted for review. We derive four requirements for any pathway that admits submissions beyond the cap and show that they can be met jointly, using identity-bound waiting as an existence proof whose ethical costs call for better implementations. In an uncertainty analysis under assumed parameter ranges, the design postpones capacity exceedance by a median of three years relative to supervision enhancement alone; the advantage is lost in most draws if its effect on load decays within about six years. Flow-by-Flow is therefore a mitigation that buys time for institutional adaptation, not a permanent remedy.

## 1 Introduction

Human review of AI outputs is a load-bearing premise of contemporary AI governance. Article 14 of the EU AI Act requires that high-risk systems be overseen by natural persons able to monitor their operation and to detect and address anomalies [1], the NIST AI Risk Management Framework asks organizations to define and document human oversight processes [2], and deployment guidelines commonly require that a qualified person check an output before it takes effect. Each of these provisions presupposes that reviewers retain the capacity to scrutinize what they approve, however much output there is to approve.

Doubts about this premise have a long history. Bainbridge [3] observed that automation leaves human operators with the tasks least suited to human cognition, namely monitoring a system that usually works and intervening when it fails in unfamiliar ways. Perrow [4] showed that accidents in complex, tightly coupled systems are structural rather than anomalous, and Beck [5] argued that technological development generates risks that exceed institutional capacities for control. Research on human-automation interaction documented complacency and automation bias [6-8], and Elish [9] described how a human placed in the loop can become a "moral crumple zone" who absorbs responsibility without retaining meaningful control. Recent legal and policy scholarship has questioned whether human-oversight mandates deliver the protection they promise [10-13]. Generative AI sharpens these concerns. Text that is easy to process tends to be judged true [14], and generative models produce fluent text whether or not it is accurate; trust in AI is a major driver of automation bias [15], explanation-based interventions have often failed to reduce it [16], and the oversight requirement of the EU AI Act has been argued to be unable to resolve it for generative models [17].

Generative AI adds a quantitative problem to these qualitative concerns. In fields where anyone may submit, the volume of material requiring accountable review is rising. Among more than two million preprints, authors who adopted large language models increased their output by 23.7% to 89.3%, depending on field and author background [18]. The volume of indexed scientific articles had already grown faster than the number of active researchers before generative AI became widespread [19]. In 2025, the computer-science section of arXiv stopped accepting review and position papers that had not passed peer review, citing a surge of submissions attributed largely to LLM-assisted writing [20], and in 2026 arXiv limited each submitter to two submissions per month [21]. An estimated 21% of the reviews submitted to ICLR 2026 were fully AI-generated [22], and about one in twenty papers accepted at NeurIPS and USENIX Security in 2025 contained at least two likely hallucinated references that had passed peer review [23]. In December 2025, patent applications filed with the Japan Patent Office reached 82,188, about 2.7 times the previous month, a surge that trade press linked to AI-assisted drafting [24].

These observations are consistent with an argument that oversight of this kind fails under stated conditions [25]. When accountability for detected errors falls on human reviewers or deploying organizations, when review capacity is bounded and grows slowly, and when an open population of submitters can expand output at low cost, review becomes nominal once the volume of material persistently exceeds capacity. The remedy suggested by this diagnosis is to limit the flow of material into review, so that effective inflow does not exceed capacity.

The diagnosis, however, leaves an institutional design problem unsolved. First, the number of submissions is a poor measure of supervisory load: a patent application with three claims and a ten-page specification and one with fifty claims, a 120-page specification and eighty cited references impose very different burdens on an examiner. Second, any mechanism that limits flow must decide which submissions to admit, and a mechanism that makes this decision by judging content faces a dilemma. If the judgment is delegated to AI, it inherits the hallucination risk it was meant to contain; if it is delegated to humans, it consumes the review capacity it was meant to protect. This paper addresses the problem of limiting supervisory load without judging content. Table 1 states its four research questions and the contributions that answer them.

**Table 1** Research questions, contributions and the sections in which they are addressed

| Research question | Contribution | Section |
|---|---|---|
| **RQ1.** Under what conditions does supervision enhancement alone fail to keep supervisory load within review capacity, and which interventions only postpone this failure rather than change its rate? | We **formulate** supervisory load as the product of submission volume $V$ and per-item load $L$, **decompose** $L$ into triage, judgment and response, and **derive** a proposition on the finite lifetime of supervision-only regimes, with corollaries that distinguish interventions on the level of load from those on its growth rate. | 2 |
| **RQ2.** What requirements must an institutional mechanism satisfy to keep supervisory load within capacity without judging the content of submissions? | We **design** Flow-by-Flow, consisting of a counting-based cognitive cost score, an institutional capacity cap and an exceedance pathway, and **derive** four requirements that any exceedance pathway must meet. | 3 |
| **RQ3.** Can these requirements be satisfied jointly, and at what practical and ethical cost? | We **analyze** which properties the design guarantees and which it does not, including the stability of the review queue and the conditions under which friction deters, and **examine** its ethical costs: proportionality, errors in load estimation, privacy, due process, timely access and unequal burdens. | 3.5, 5, 7 |
| **RQ4.** Under what parameter conditions does such a mechanism buy more time than supervision enhancement alone? | We **evaluate** the design through an uncertainty analysis with break-even and sensitivity analyses. | 6 |

The contribution is theoretical. The constraint $V \times L \leq C$ introduced below is an uncalibrated hypothesis about institutional capacity, the cognitive cost score is a proposed proxy rather than a validated measure of cognitive load, and the simulation propagates assumed parameter ranges rather than estimating them.

The analysis applies where four conditions hold jointly: errors can cause severe or irreversible harm, directly or through institutions that rely on the outputs (high loss); correctness cannot be established by a mechanical or formal check; accountability for detected errors falls on a human reviewer or a deploying organization; and the population of submitters is open, so that output can grow faster than review capacity. Patent examination, court filings and academic peer review are examples. Peer review counts as high-loss because clinical guidelines, regulatory decisions and public policy rely on the scientific record: in 2020, a paper based on data that could not be verified led the World Health Organization to pause the hydroxychloroquine arm of a major trial before the paper was retracted [26]. AI-assisted diagnosis by a fixed medical staff does not meet the last condition: volume follows a caseload that the institution controls, and supervision enhancement may keep pace. Domains whose outputs can be verified mechanically fall outside the scope. The scope is therefore determined by the structure of a domain, not by its name.

Table 2 summarizes seven standard objections to limiting flow and our short responses. It is intended as a practical checklist for evaluating oversight proposals: for each objection, a policymaker can ask whether a proposed safeguard changes the growth rate of supervisory load or only its level, which Section 2.4 shows to be the decisive distinction. Appendix A develops the responses.

**Table 2** Standard objections to limiting flow into accountable review, and short responses (developed in Appendix A)

| Objection | Response |
|---|---|
| 1. If AI becomes more accurate, supervision will become easier. | Higher accuracy reduces the error probability $p$ but does not make it zero. It can also increase volume and the number of accepted outputs that require a downstream response. |
| 2. AI can verify AI. | Verification by a probabilistic system must itself be verified. Where accountability rests with a person, approval based on an unverified verifier certifies what has not been reviewed. |
| 3. More reviewers can be hired. | Hiring raises capacity linearly at best, whereas supervisory load can grow at a compound rate (Section 2.4). |
| 4. Disclosure of AI use is enough. | Disclosure that cannot be verified at scale does not remove the need to review every submission as potentially AI-generated. |
| 5. Full intake is an access right. | The right to submit does not imply a right to immediate processing; other infrastructures distinguish access from simultaneous throughput (Section 3.2). |
| 6. Well-resourced actors can bypass friction. | They can, by paying for it. The design aims to attach a price to the consumption of review capacity, not to exclude anyone (Sections 5.1 and 5.3). |
| 7. Nominal oversight is acceptable if AI errs less than humans. | In high-loss domains tolerance is set by the absolute number of consequential failures. If volume grows faster than the error rate falls, failures increase. |

The paper is organized as follows. Section 2 develops the theoretical framework and derives Proposition 1. Section 3 presents the Flow-by-Flow design and the requirements for its exceedance pathway. Section 4 uses a stylized model of applicant behaviour to explain why the cost score combines several features. Section 5 shows with identity-bound waiting that the requirements can be met jointly and analyzes when friction deters. Section 6 reports the uncertainty analysis. Section 7 examines the ethical costs of the design, Section 8 states its limitations, and Section 9 concludes.

## 2 Theoretical Framework: From $V$ to $V \times L$

This section restates the throughput argument in a self-contained form, decomposes per-item load, and derives the dynamic result on which the rest of the paper relies. Table 3 lists the symbols.

**Table 3** Symbols

| Symbol | Definition |
|---|---|
| $V$ | Number of outputs submitted for accountable review per period |
| $L$ | Per-item load: reviewer time required to review one output at the required depth |
| $C$ | Review capacity: reviewer time available per period at the required depth ($C_{\text{eff}} \leq C_{\text{max}}$) |
| $W$ | Supervisory load, $W = V \times L$ |
| $g_V, g_L, g$ | Annual growth rates of $V$, $L$ and $W$, with $1 + g = (1 + g_V)(1 + g_L)$ |
| $a$, $b$ | Linear and compound growth rates of capacity |
| $s$ | Initial slack, $s = C_0/W_0$ |
| $t^*$ | Time at which $W$ first exceeds $C$ |
| R | Attribution of accountability; $R = 1$ when it rests with a natural or legal person rather than the AI system |
| $p$ | Probability that an output contains an error, $p > 0$ |
| $S$ | Cognitive cost score computed from countable features $x_i$ with weights $w_i$ |
| $k$, $T$ | Exceedance multiplier of the score and base waiting time of the exceedance pathway |
| $N$ | Number of features in the score |

### 2.1 The throughput constraint

Let $V$ be the number of outputs that require accountable review in a period, $L$ the reviewer time needed to review one output at the depth the institution requires, and $C$ the reviewer time available per period at that depth. Substantive review is possible only if

$$V \times L \leq C. \qquad (1)$$

The upper bound $C_{\max}$ reflects well-established limits on attention, working memory and sustained vigilance [27-29]; the effective capacity $C_{\text{eff}}$ may fall below it under fatigue or overload. We write $C$ for the effective capacity of a reviewing organization, which training, tools and hiring can raise. When (1) is violated, the coverage ratio $C/(VL)$ falls below one: either some outputs go unreviewed or each is reviewed less thoroughly, and in both cases the probability $d$ that an error is detected falls. With error probability $p > 0$, the number of undetected errors per period, $Vp(1-d)$, then rises with volume, while the errors that institutions observe, and to which their feedback responds, number only $Vpd$ [25].

Evidence from patent examination links the time available per item to the quality of review: promotions that reduce the hours allocated to each application are followed by less search, more grants and grants of lower average validity [30]. Constraint (1) generalizes this observation: when review time per item is squeezed, review becomes more lenient and less reliable.

Three conditions make violations of (1) persistent rather than transient. First, accountability for detected errors rests with human reviewers or deploying organizations ($R = 1$), so the duty to review cannot be discharged by pointing to the AI system. Second, capacity is bounded and grows slowly: training, tools and hiring raise $C$, but not at the rate at which AI-assisted output can grow. Third, the population of submitters is open, so nothing ties the volume of submissions to what reviewers can check. Under these conditions oversight becomes nominal once (1) is persistently violated [25]. The present paper takes these conditions as premises; the evidence for the third is reviewed in Section 1, and Section 2.4 shows that only a weak form of the second is required.

## 2.2 Components of per-item load

The number of outputs alone does not measure supervisory load. A reviewer of a thermometer reading knows in advance that the reading is a temperature and compares it with an external standard. A reviewer of a generative-AI output does not know in advance what kind of statement it contains. We therefore distinguish three components of per-item load that differ in kind. *Triage* is the task of classifying the semantic type of an output, that is, deciding whether a passage is a factual report, an inference, an opinion, a narrative or a calculation, and hence which standard of evaluation applies: truth for facts, plausibility for inferences, coherence for narratives, correctness for calculations and justification for opinions. *Judgment* is the evaluation of the content under that standard. *Response* is the institutional action that follows judgment, such as drafting a reasoned rejection, revising a treatment plan or writing a software patch. This definition of triage, which differs from the medical sense of ordering cases by urgency, is used throughout the paper.

The three components respond differently to improvements in model accuracy. Judgment cost is under pressure to fall, but largely through omission: when errors are rare, reviewers check less, and rare errors are disproportionately missed [8, 31]. Response cost does not fall with accuracy, because it scales with the number of accepted outputs and depends on human labour; a more accurate model that produces more accepted outputs creates more responses. Triage cost does not fall either, because generality is the defining property of general-purpose models: their outputs are not confined to a fixed semantic category in the way that the outputs of a thermometer or a calculator are.

Whether per-item load grows over time matters for the dynamics of Section 2.4, and one possible source of such growth is uncertainty about the provenance of an output. This uncertainty is distinct from triage, but it can raise triage cost. When a reviewer can attribute a statement to a known source, the source carries information about how the statement was produced, for example whether it reports a measurement, an expert's considered judgment or a speculation, and thus about which standard should apply. When provenance is uncertain, this information is missing, and the semantic type must be inferred from the text alone. Experiments show that information about origin enters evaluation: labelling the same content as AI-generated lowers its perceived accuracy, regardless of whether it is true or who wrote it [32, 33]. Human ability to tell AI-generated from human-written text is limited: people rely on flawed heuristics [34], and although some individuals perform above chance, accuracy is modest and varies widely [35]. We therefore expect provenance uncertainty to raise triage cost, but the size of this effect has not been measured. Section 6 therefore treats the growth of $L$ as uncertain and reports results for the case in which $L$ does not grow at all.

## 2.3 Growth of load and capacity

Let volume and per-item load grow at annual rates $g_V$ and $g_L$, so that supervisory load $W = V \times L$ grows at rate $g$:

$$W(t) = W_0(1+g)^t, \qquad 1+g = (1+g_V)(1+g_L). \qquad (2)$$

Supervision enhancement, such as hiring and training reviewers, is represented by capacity that grows linearly,

$$C(t) = C_0(1+at), \qquad a \geq 0, \qquad (3)$$

or, as an alternative, at a compound rate $h$. The evidence in Section 1 supports $g_V > 0$ for open populations of submitters. The sign and size of $g_L$ are less certain. The argument below requires only that $g > 0$ under (3), or $g > h$ under compound capacity growth.

## 2.4 Proposition 1: the finite lifetime of supervision-only regimes

**Proposition 1.** Let load follow (2) with $g > 0$, let capacity follow (3), and let the initial slack be $s = C_0/W_0 > 1$. Then load exceeds capacity in finite time. The first exceedance time $t^*$ is the unique positive solution of $(1 + g)^{t^*} = s(1 + at^*)$, that is,

$$t^* = \frac{\ln(s) + \ln(1 + at^*)}{\ln(1 + g)}. \qquad (4)$$

*Proof.* Let $f(t) = t\ln(1 + g) - \ln(s) - \ln(1 + at)$, so that $W(t) > C(t)$ if and only if $f(t) > 0$. Then $f(0) = -\ln(s) < 0$, and $f(t) \to \infty$ as $t \to \infty$ because $t\ln(1 + g)$ grows linearly while $\ln(1 + at)$ grows logarithmically. Since $f$ is continuous, it has a positive root. Since $f''(t) = a^2/(1 + at)^2 \geq 0$, $f$ is convex, and a convex function that is negative at zero and diverges to infinity has exactly one positive root. Rearranging $f(t^*) = 0$ gives (4). ▫

The initial slack $s$ and the rate of capacity growth $a$ enter (4) only through logarithms, whereas the growth rate of load enters through the denominator. Three corollaries make this asymmetry precise.

**Corollary 1 (level interventions).** Multiplying initial capacity by $\kappa > 1$, or reducing load permanently by a fraction $\sigma \in (0,1)$, is equivalent to replacing $s$ with $\kappa s$ or $s/(1 - \sigma)$. When $a = 0$, it postpones exceedance by exactly $\ln\kappa/\ln(1 + g)$ or $\ln(1/(1 - \sigma))/\ln(1 + g)$ years; when $a > 0$, the postponement is at least as large and of the same logarithmic order. At $g = 0.30$, for example, doubling capacity postpones exceedance by 2.6 years, and a permanent 95% reduction of load postpones it by 11.4 years. A reduction that fades over time postpones it by less.

**Corollary 2 (rate interventions).** Suppose that from time $t_0 < t^*$ the growth rate of load falls from $g$ to $g'$ with $0 < g' < g$. When $a = 0$, the remaining time to exceedance, $t^* - t_0$, is multiplied by $\ln(1 + g)/\ln(1 + g')$; halving a moderate growth rate roughly doubles it. If $g' \leq 0$, load does not exceed capacity after $t_0$. A limit on the number of submissions that binds from $t_0$ sets $g_V$ to zero thereafter and so replaces $g$ with $g_L$.

**Corollary 3 (compound capacity growth).** If capacity grows as $C(t) = C_0(1 + h)^t$, load exceeds capacity in finite time if and only if $g > h$, at $t^* = \ln(s)/[\ln(1 + g) - \ln(1 + h)]$.

*Proof of Corollaries 1 to 3.* Corollary 1 follows from substituting $\kappa s$ or $s/(1 - \sigma)$ for $s$ in (4); for $a > 0$ the right-hand side of (4) increases with $t^*$, so the new solution exceeds the old one by at least the change in $\ln(s)/\ln(1 + g)$. For Corollary 2 with $a = 0$, load at $t_0$ is $W_0(1 + g)^{t_0}$ and grows at rate $g'$ thereafter, so exceedance occurs at $t_0 + [\ln(s) - t_0\ln(1 + g)]/\ln(1 + g')$, whereas without the intervention it occurs at $t_0 + [\ln(s) - t_0\ln(1+g)]/\ln(1+g)$. Corollary 3 follows from solving $W_0(1+g)^t = C_0(1+h)^t$. ▫

The institutional reading is direct. Hiring and training reviewers, giving them better tools, and AI assistance that raises their throughput by a bounded factor act on the level of capacity; one-off reductions of load, such as a stricter threshold or a deterrent whose effect fades, act on the level of load. All of these buy time only logarithmically. Interventions that change the growth rate of the load admitted to review, such as per-submitter limits on the number of submissions or a cap on admitted load, act on the denominator of (4). Supervision enhancement and flow limitation are therefore not interchangeable remedies of the same kind. Supervision enhancement prevents exceedance only if capacity can be made to grow at the same compound rate as load (Corollary 3), which for capacity based on trained human reviewers would require the reviewing workforce to grow as fast as AI-assisted output.

The mathematics of Proposition 1 is elementary. Its relevance lies in the fact that the oversight mandates cited in Section 1 specify what reviewers must do but not how review capacity is to keep pace with compound growth in the volume they must review.

### 2.5 From diagnosis to design

The responses most often proposed to the problem of overloaded review are more accurate models, more reviewers, AI-based verification of AI outputs, and mandatory disclosure of AI use. Each may be useful, but none bounds the growth rate of supervisory load. Accuracy improvements reduce $p$ but can increase $V$ and the number of downstream responses; additional reviewers raise the level of capacity; AI verification of AI outputs must itself be verified when accountability rests with a person; and disclosure that cannot be verified does not remove the need to review each output on the assumption that AI may have produced it (Table 2; Appendix A). Flow-by-Flow, presented next, does not judge whether outputs are correct. It controls how much supervisory load is admitted to review, so that judgment by accountable humans remains possible within their capacity.

## 3 The Flow-by-Flow Design

### 3.1 Design principle

The aim of Flow-by-Flow is to prevent outputs containing hallucinations, factual distortions or unwarranted inferences from passing through review in greater volume than accountable reviewers can scrutinize. The design pursues this aim without judging whether any output is correct. It rests on a single principle: *to implement, without content judgment, incentives that make high-volume and high-complexity submission costly while keeping concise submission inexpensive.* The principle requires no judgment about intent. It is unnecessary to distinguish good-faith from malicious use of AI, or legitimate from illegitimate high-volume submitters; what the design prices is the consumption of finite review capacity.

The design has three components. A *cognitive cost score* estimates the supervisory load of each submission from countable features (Section 3.3). An *institutional capacity cap* fixes the load admitted to review in each period (Section 3.5). An *exceedance pathway* admits submissions that exceed the score threshold or the submitter's allowance, at an identity-bound cost per submission (Sections 3.5 and 3.6). For a typical submitter, all three are inexpensive: a submission within ordinary bounds passes the first component automatically and is not affected by the third. For a submitter who multiplies the number or the complexity of submissions, the cost grows with the load imposed.

### 3.2 Access and immediate processing

Knowledge infrastructures in high-loss domains, such as patent offices, courts, peer-review systems and drug-approval agencies, implicitly rely on a syllogism: (a) these infrastructures are important; (b) everyone has a right of access to important infrastructures; (c) therefore every submission must be processed as soon as possible. Premises (a) and (b) are sound, but (c) does not follow from them. The right of access guarantees that anyone may submit; it does not guarantee immediate processing.

Other infrastructures draw this distinction routinely. Hospitals treat all patients but order treatment by severity; roads admit all vehicles but meter simultaneous passage with signals; electricity grids serve all consumers but manage peak demand; communication networks connect all users but allocate bandwidth. In each case the obligation to serve everyone is distinguished from the obligation to serve everyone at once. Queues already allocate access implicitly whenever demand exceeds capacity, and economists have long analysed waiting time as an allocation mechanism with its own efficiency and distributional properties [36, 37]. Under overload, access to review is rationed in either case, and the design choice is whether to ration explicitly, under stated rules, or implicitly, through delay and declining scrutiny.

Before generative AI, the effort needed to draft a patent application or a paper limited the rate of submission, so that volume seldom exceeded the capacity of the reviewing institution, and the practice of processing every submission promptly could persist without being chosen as a principle. Generative AI removes this natural limit. Maintaining full and immediate processing under these conditions does not protect access; it degrades review for everyone, and a legitimate submitter whose application receives only nominal review has in effect been denied the examination the institution exists to provide. Flow-by-Flow keeps the right to submit intact and secures the legal date of each submission at the moment of filing (Section 7.5), while making the order and conditions of processing explicit.

### 3.3 The cognitive cost score

The first component estimates supervisory load without evaluating content. It uses only countable features: in patent applications, the number of claims, the length of the specification and the number of cited references; in academic manuscripts, word count, number of references, number of figures and tables, and volume of supplementary material; in court filings, length, number of evidentiary items and number of provisions invoked; in drug applications, the length of the dossier, the number of clinical trials and the number of indications. Counting requires no semantic judgment. A counting error, such as recording 62 claims instead of 61, is verifiable and mechanically correctable, unlike a hallucinated judgment that a claim is novel.

Institutions already treat such features as proxies for examination burden. The European Patent Office charges a fee for each claim beyond the fifteenth and for each page beyond the thirty-fifth [38]; the United States Patent and Trademark Office charges for claims beyond twenty, for independent claims beyond three, and for application size beyond 100 sheets [39]; and the Japan Patent Office's request-for-examination fee includes a per-claim component [40]. Applicants respond to such pricing: after the EPO introduced excess-claims fees in 2008, claim counts fell while the breadth of protection was maintained [41]; where claims are priced but specification length effectively is not, specifications have lengthened for two decades [42]; and practitioners recommend consolidating claims to avoid claim fees [43]. These responses illustrate both the premise of the score, that formal features track burden closely enough to be priced, and its main vulnerability, that single features invite substitution (Section 3.4).

Evidence on the duration of examination points the same way. In a study of more than 200,000 applications to the European Patent Office, the number of claims, the feature most directly under the applicant's control, lengthened examination whatever its outcome, and examination also took longer when the office's workload at the time of filing was higher [44]. The United States Patent and Trademark Office went further in 2007, when it adopted a rule limiting applicants to five independent and twenty-five total claims as of right [45]. The rule was challenged in court and rescinded in 2009 [46], which also illustrates the legal obstacles that limits of this kind face (Section 7.5).

For each feature $i = 1, \dots, N$, let $f_i$ be its count for a submission and $m_i$ the median count in a fixed reference set of past submissions in the same field. The normalized feature is $x_i = f_i / m_i$, and the score is

$$S = \prod_{i=1}^{N} \max(1, x_i)^{w_i}, \qquad w_i > 0, \qquad (5)$$

or equivalently $\ln S = \sum_i w_i \ln^+ x_i$, where $\ln^+ x = \max(0, \ln x)$. Normalization by the median makes features with different units comparable. Because of the floor at one, features at or below the reference median contribute nothing, so a submission at or below the median in every feature has $S = 1$.

Table 4 compares (5) with three alternatives, using the example of two submissions with normalized features $A = (2,2,2)$ and $B = (10,0.2,0.2)$. An unfloored product lets a small value in one dimension offset a large value in another, so that $B$, which is ten times the median in one dimension, scores far below $A$. The floored product (5) shares four properties: it is non-decreasing in every feature; a below-median feature cannot offset an above-median one; simultaneous excess in several features compounds; and it equals one for ordinary submissions, so that they bear no friction.

**Table 4** Candidate functional forms for the cognitive cost score (unit weights)

| Form | Below-median features offset excess | Simultaneous excess compounds | $S(A)$ | $S(B)$ | Natural calibration model |
|---|---|---|---|---|---|
| Weighted sum $\sum_i w_i x_i$ | Partly | No | 6 | 10.4 | Linear regression of review time on features |
| Product $\prod_i x_i^{w_i}$ | Yes, without bound | Yes | 8 | 0.4 | Log-linear regression on $\ln x_i$ |
| Floored product (5) | No | Yes | 8 | 10 | Log-linear regression on $\ln^+ x_i$ |
| Maximum $\max_i x_i$ | No | No | 2 | 10 | None |

Three considerations favour (5). First, review work often involves interactions between features: an examiner compares each claim with each cited reference, so the work of comparison grows with the product of the two counts rather than their sum. Second, the form is testable. Each institution can regress the logarithm of recorded review time on $\ln^+ x_i$, estimate the weights $w_i$, and compare the fit out of sample with that of the additive alternative; where review time turns out to be additive in the features, the additive form should be adopted. Third, compounding makes it costly to inflate several dimensions at once, which is the strategy that single-feature pricing invites. None of these considerations establishes that $S$ measures cognitive load. The score as a whole is a proposed proxy whose relationship to review time must be estimated in each institution before use (Section 8).

These considerations also explain why a product of counts is a reasonable starting point despite being crude. Individual features such as the number of claims are known to lengthen examination, but no composite score of this kind has yet been validated against actual review time in any of the domains considered here, so the first formulation should be the one that existing records can test and correct most quickly. In logarithms the floored product is linear in the features, so its weights and its fit relative to the additive alternative can be estimated by ordinary regression on recorded review times, and a poor fit becomes visible as soon as such data are examined. Contract theory offers an analogous result: in a principal-agent model in which contracts are evaluated by their worst-case performance over actions unknown to the principal, simple linear contracts are optimal [47], which suggests that simple rules can be robust when a designer knows little about the environment. Finally, the protection of review does not depend on the precision of the score. The capacity cap limits the time that reviewers actually spend, so it bounds reviewer load whatever the error of the score (Section 3.5 and Proposition 2), and the durable part of the time that the design buys comes from limits on submissions rather than from the score (Finding 3 in Section 6). If no count-based score explains review time adequately in a field, setting $S = 1$ for all submissions reduces the design to limits on the number of submissions together with the cap on review time, and both properties are retained. A crude score therefore costs accuracy in routing, which calibration and the reconciliation of Section 7.3 can improve, but it does not cost the protection of review.

If the reference medians $m_i$ were updated continuously, widespread use of AI could shift them upwards, and the score would gradually cease to flag the submissions that impose the most load. The reference set is therefore fixed to a stated period, ideally one preceding widespread use of generative AI, and only the weights $w_i$ are recalibrated against review-time data. A population-wide rise in complexity then raises the scores of all submissions and is absorbed by the capacity cap (Section 3.5) rather than by renormalization. If legitimate practice in a field changes, the reference set can be revised by an explicit, scheduled and published decision, with a limit on the size of each revision, rather than by automatic updating.

### 3.4 Evasion through optimization

Any published metric invites optimization against it [48]. If submission counts are limited, submitters increase length and complexity; if length is limited, they compress and increase density. Before generative AI, discovering and implementing such strategies required human effort, which itself acted as friction. Constraint-satisfying rewriting, such as "compress this specification to fifteen pages while preserving its scope," is now an ordinary task for large language models, a capability that rests on the Transformer architecture, on capability scaling and on instruction tuning [49-51]. Single-feature limits therefore become more fragile as models improve.

The score combines several features for this reason. Because each feature is tied to the substance of the submission, compressing one dimension transfers burden to another: fewer claims require more description per claim, and a shorter text requires denser argument. Evading several mutually constraining features at once is a multidimensional constraint-satisfaction problem, harder than evading one. This robustness is relative, not absolute. As optimization capability grows, every countable dimension can eventually be saturated, at which point remaining differences between submissions lie in uncountable qualities, such as the depth of an argument, that only content judgment can assess. Flow-by-Flow does not remove this endpoint; it postpones it. Section 6 treats the rate at which the score's effect decays as one of the two most uncertain parameters of the analysis.

### 3.5 The capacity cap, the exceedance pathway and queue stability

The second component makes the reviewing institution's capacity an explicit rule. The cap $K$ limits the total time that reviewers with authority to decide spend on review in each period; an institution may, for example, limit the combined review time of its examiners to 100 hours a week. The cap and the depth of review it must allow are set not from cognitive measurements of individuals but from operational records: the relationship between daily workload and error rates (missed defects, incorrect approvals, later reversals) is plotted for each task type, and the workload is held conservatively below the point at which error rates begin to rise, with periodic audits using seeded errors to detect later degradation (Section 8).

Submissions are processed along two paths. A submission by an identity-verified submitter that is within the submitter's allowance $\bar{n}$ per period and has $S \leq S_0$ enters the review queue directly. Any other submission enters the review queue through the *exceedance pathway*, which attaches a cost to that individual submission, scaled by the exceedance multiplier $k = \max(1, S/S_0)$. Section 5 presents one implementation. Reviewers take submissions from the queue in order of entry. A submission counts as completed only when it has been reviewed to the required depth, and work not finished when the period's review time is used up is carried over to the next period. Because the cap applies to the time actually spent, it holds even when the score underestimates the load of a submission: a heavier submission takes more of the 100 hours and fewer submissions are completed that week, but review time does not exceed 100 hours. The first layer of the design, measurement and routing, is fully automatic and involves no content judgment; the second layer, review, is human and receives only what capacity allows.

This architecture guarantees one property and not another.

**Proposition 2 (what the cap guarantees).** Let $A_t$ be the review time actually required by the submissions admitted to the review queue in period $t$ and $Q_t$ the review time still required by the backlog, with $Q_{t+1} = \max(0, Q_t + A_t - K)$. (i) Review time never exceeds $K$ in any period, whatever the arrivals. (ii) If admitted load exceeds $K$ by at most $B$ over any run of consecutive periods, that is, $\sum_{\tau=u}^{t-1}(A_\tau - K) \le B$ for all $u < t$, the backlog never exceeds $Q_0 + B$; if the long-run average of $A_t$ exceeds $K$, the backlog grows without bound. (iii) If unrationed demand $W_t$ grows at a compound rate $g > 0$, a bounded backlog requires the share of demand that is admitted, $A_t/W_t$, to fall towards zero at rate $g$, in proportion to $K/W_t$.

*Proof.* Part (i) holds by construction. For part (ii), unrolling the recursion gives $Q_t = \max\{Q_0 + \sum_{\tau=0}^{t-1}(A_\tau - K),\ \max_{0<u<t} \sum_{\tau=u}^{t-1}(A_\tau - K),\ 0\}$, which is at most $Q_0 + B$ under the stated condition and diverges if the average of $A_\tau - K$ is positive. For stochastic arrivals, the standard stability condition of queueing theory, a traffic intensity below one, plays the same role. Part (iii) follows from part (ii) with $W_t = W_0(1+g)^t$. ▫

Because $A_t$ and $Q_t$ are measured in the review time that submissions actually require, part (i) holds whatever the error of the score, provided that the cap is applied to the time reviewers spend, as described above. If an institution instead admitted submissions until their estimated load reached $K$ and required all of them to be completed within the period, underestimation would push review time above $K$, and the cap would need a margin below capacity sized to the error of the score. Applied to review time, the cap guarantees bounded reviewer load, which is the property that keeps review substantive. It does not by itself guarantee a bounded backlog. By Little's law, the expected time from admission to review equals the backlog divided by throughput [52], so an unbounded backlog means unbounded delay in obtaining a decision. A bounded backlog requires that admission itself be regulated: the base cost $T$ of the exceedance pathway, the threshold $S_0$ and the allowance $\bar{n}$ can be adjusted so that admitted load does not exceed $K$, in the manner of a congestion price [53]. If the friction needed to clear demand would exceed what is proportionate (Section 7.2), the institution must ration by other means, such as quotas per submitter or lotteries among submissions with protected legal dates, or it must accept a growing backlog. Part (iii) shows that under compound growth of demand this choice cannot be avoided, only postponed. The exceedance pathway has its own capacity: if identity verification can be performed $\mu_X$ times per period, the number of submissions using the pathway must stay below $\mu_X$, or a second queue forms in front of it.

What the design adds is that overload is converted from an invisible collapse of review quality into an explicit allocation problem, which can be governed, audited and contested (Section 7).

### 3.6 Requirements for the exceedance pathway

The score and the cap involve no content judgment and do not consume reviewer time, but neither governs the submissions that exceed them. An exceedance pathway must admit such submissions without judging whether they are correct and without using reviewer time, while preventing the near-zero cost of replicating submissions with AI from carrying over into the pathway. The first two are constraints on the institution; the third is a constraint on the cost borne by submitters. We derive four requirements from this purpose under a threat model with four features: (T1) submissions can be generated and replicated at near-zero marginal cost; (T2) identities can be created or borrowed unless they are verified; (T3) any component that judges content either can be optimized against and can err (if automated) or consumes reviewer capacity (if human); and (T4) wealth differs across submitters by orders of magnitude, whereas the time available to each person does not. We call the requirements design invariants.

1. **No content judgment.** The pathway must not evaluate the truth, quality or appropriateness of a submission. If it did, it would reintroduce hallucination risk and an optimization target when automated (T3), or the capacity constraint when performed by humans.
2. **No scalable consumption of reviewer capacity.** The pathway must not consume reviewer time in proportion to the number of submissions that use it. If it did, the pathway itself would become subject to the constraint (1) it is meant to protect.
3. **Identity-bound friction.** The cost of each exceedance submission must be paid in the time of a verified person, not in a transferable resource such as money. Without identity binding, a submitter could avoid the pathway altogether by multiplying its allowance across fabricated or borrowed accounts (T2). If the cost could be paid in money, a fee high enough to matter to a well-capitalized submitter would exclude many others (T4). Delegation does not breach this requirement: a submitter may engage others to wait, but each of them spends their own verified time, so the volume passing through the pathway is bounded by the number of people mobilized rather than by capital (Section 5.3).
4. **No batch clearance.** The cost must attach to each submission separately, so that every additional submission carries its own cost: no single action, credential or certification, and no payment that stays the same as the number of submissions grows, may clear more than one submission. Any such batch route restores the near-zero marginal cost of replication (T1).

Each invariant is necessary in the following sense: if it is violated, there is a class of strategies that reduces the marginal cost of an additional exceedance submission to near zero, returns the load to reviewers, or lets capital rather than verified human time set the limit on what passes through the pathway. Violating Invariant 1 creates a target for adversarial optimization or a human bottleneck; violating Invariant 2 makes the pathway a bottleneck of the same kind; violating Invariant 3 lets a submitter escape the pathway by multiplying accounts or pass through it by payment, so that the limit is set by capital; and violating Invariant 4 permits clearance in bulk. This is a structured argument under the stated threat model, not a formal proof, and we do not claim that the four invariants are jointly sufficient. Fees charged per claim or per page satisfy

Invariants 1, 2 and 4, since every additional submission still costs its fee, but not Invariant 3, because a fee is paid in money rather than in the time of a verified person (Appendix B.2). The cost score and the capacity cap satisfy Invariants 1 and 2. Section 5 presents a mechanism that satisfies all four; any mechanism that does so could replace it.

# 4 A Stylized Model of Applicant Behaviour

This section uses a stylized model of applicant behaviour to explain why the cost score combines several features and why a limit on the number of submissions is not sufficient on its own.

## 4.1 Two stages of control

Control of AI-assisted submission in high-loss domains can be introduced in two stages. The first stage combines identity verification with a limit on the number of submissions per verified submitter and period. The two elements are inseparable: a limit without verification is evaded by creating additional accounts, and verification without a limit does not constrain volume. Outside high-loss domains, the preprint server arXiv introduced such a control in 2026, limiting each submitter to two submissions per month [21]. Patent offices and courts verify the identity of applicants but do not limit the number of filings, and most peer-review systems neither verify identity effectively nor limit submissions. The second stage adds the cognitive cost score and the exceedance pathway of Section 3.

## 4.2 Applicant behaviour under a limit on submissions

Applicants seek favourable outcomes, such as granted patents or accepted papers, and weigh the expected return from a submission against the costs of preparing it and of passing institutional controls. Three types can be distinguished. Applicants who do not use AI face high preparation costs per submission, so the number and complexity of their submissions are limited by their own time. Applicants who use AI face much lower preparation costs and form a continuum: at one end they use AI to draft more efficiently what they would have submitted anyway; at the other they aim to maximize the number of submissions and generate content that is likely to pass. The model does not need to locate individuals on this continuum: it concerns only the distribution of features that the population of applicants produces under a given rule, whatever the motives of individual applicants.

Under a limit on the number of submissions, the only remaining margin of optimization is the content of each submission. Because the expected return of a patent application or a paper rises with its scope and comprehensiveness, applicants have an incentive to add claims, pages and references. For applicants who do not use AI this incentive is checked by the cost of writing; for applicants who use AI it is checked only weakly. A first-stage control can therefore hold $V$ constant while $L$ rises, so that $V \times L$ exceeds capacity even though the number of submissions appears to be under control. This outcome requires no intent to evade the rules: it is the aggregate result of individually reasonable responses to the rule. Its speed differs between domains. Where much of the work of preparing a submission takes place in physical space, as in clinical trials or fieldwork-based research, each submission contains elements that AI cannot generate, and the shift towards denser submissions is slower. Where the work takes place almost entirely in information space, as in drafting patent specifications, academic manuscripts or legal briefs, the shift can follow a limit on submissions almost immediately.

### 4.3 A second stage based on several features

The second stage adds the cognitive cost score to the first. Consider patents with three normalized features: text length $X$, number of claims $Y$ and number of cited references $Z$, so that under (5) with unit weights $S = \max(1, X)\max(1, Y)\max(1, Z)$. Because a higher score brings friction on the exceedance pathway, an applicant can no longer raise one feature freely: raising $X$ above the median requires keeping $Y$ or $Z$ at or below it if the submission is to stay below the threshold. The features are tied to the substance of the submission, so these trade-offs have substantive consequences: fewer claims narrow the scope of protection, and a shorter specification omits description. Adding a feature is cheap for the institution, whereas each additional feature adds a dimension to the applicant's optimization problem.

Figure 1 illustrates the intended dynamics. Under a limit on submissions alone, both the score and the actual load of submissions drift upwards over time. Under the second stage, the score is capped at the threshold, and actual load can rise only through dimensions that the score does not capture, which is costlier; as optimization capability improves, this suppression fades. The figure is illustrative: its growth rates were chosen for exposition and are not estimated or simulated from data.

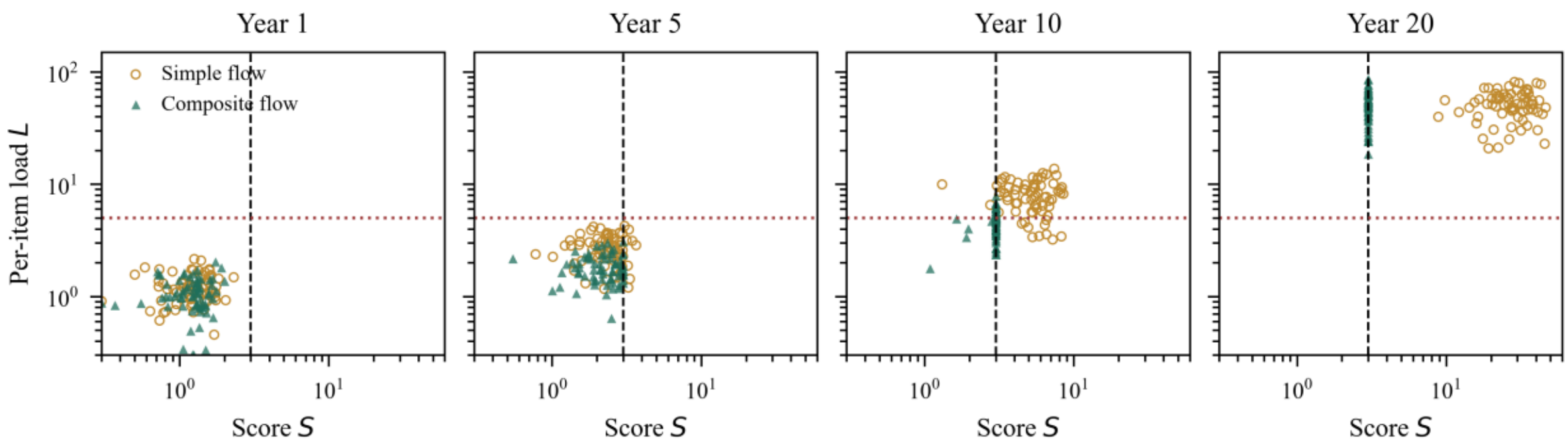

**Fig. 1** Illustrative dynamics of the cognitive cost score $S$ (horizontal axis) and actual per-item load $L$ (vertical axis) for submissions under a limit on submissions alone (simple flow, circles) and with the cost

score added (composite flow, triangles), in years 1, 5, 10 and 20. The dashed vertical line marks the score threshold $S_0$ and the dotted horizontal line the per-item load at which aggregate load reaches capacity. Both axes are logarithmic. Under simple flow, $S$ and $L$ both rise; under composite flow, $S$ is held at the threshold and $L$ rises more slowly until the suppression has decayed. Growth rates are chosen for exposition; the figure is not a simulation result

## 5 Identity-Bound Waiting as an Existence Proof

This section presents one mechanism that satisfies the four invariants of Section 3.6. Its role is only to show that the invariants can be satisfied jointly. We do not recommend it as it stands: its costs, examined in Section 7, call for better implementations.

### 5.1 Mechanism

A submission routed to the exceedance pathway requires a period of waiting at a designated office that verifies identity. The submitter presents an identity document at the start and the end of the waiting period and then receives a single-use passcode bound to that submission, which admits it to the review queue. The waiting time is proportional to the exceedance multiplier of the score:

$$\text{waiting time} = T \times k, \qquad (6)$$

where $T$ is a base time, for example two hours. A submission with $k = 8$ thus requires sixteen hours, which may be divided into separately verified sessions. $T$ is an institutional parameter that can be adjusted without changing any algorithm or judging any content.

The mechanism satisfies the four invariants. It involves no evaluation of content (Invariant 1). It consumes the time of staff who verify identity, not of reviewers (Invariant 2). Verification at the start and end of each waiting period binds the cost to the time of a verified person (Invariant 3). Each submission requires its own waiting period and passcode, so no action clears more than one submission (Invariant 4).

The mechanism is best understood as congestion pricing on review capacity [53]. A congestion charge does not ask whether a journey is legitimate; it prices the use of a scarce shared resource. Likewise, the exceedance pathway does not ask whether a submitter is legitimate. It attaches a time cost to each unit of load that exceeds ordinary bounds. For an applicant who occasionally needs a complex submission, the cost is bounded and occasional. For any high-volume submitter, human or AI-assisted, the cost grows at least linearly with the number of exceedance submissions and cannot be reduced by digital replication. A single user can generate 100,000 drafts at negligible marginal cost through an interface to a language model; under this mechanism, 100,000 exceedance submissions require 100,000 periods of identity-verified waiting.

The mechanism belongs to the lineage of proof-of-work, proposed to make unsolicited bulk email costly [54], and of tests that are easy for humans and hard for machines [55]. Both lines of defence have weakened. Machine-solvable tests are now widely circumvented [56], and computational proof-of-work fails when the cost low enough for legitimate senders is negligible for attackers with more computing power [57]. Identity-bound waiting differs in what it consumes. Its scarce input is verified human presence over time, which hardware cannot multiply: a submitter who wants $n$ exceedance submissions must supply $n \times T \times k$ hours of verified presence, either personally or by employing other people.

### 5.2 Accommodation and a remote variant

The purpose of the mechanism is not physical travel but the consumption of a finite amount of identity-verified time per exceedance submission. Regional offices, appointment systems and registered proxies who are themselves identity-verified are therefore compatible with it, provided that each submission still requires its own waiting period. A remote variant, an identity-verified time lock, verifies the submitter's identity remotely at the beginning and end of a session that must last a minimum time before the passcode is issued. The remote variant is more accessible but more exposed to delegation of the waiting to hired labour, since remote presence is harder to verify than physical presence. Which variant is appropriate in a given jurisdiction is a question of proportionality (Section 7.2).

### 5.3 Conditions for deterrence

Friction deters a submission only if it costs more than the submission is worth to the submitter. Let $\pi$ be the expected payoff of an exceedance submission, $c_0$ its preparation cost, $w$ the hourly cost of the person who waits (the submitter's own opportunity cost or the wage of an employed proxy) and $\kappa$ a fixed coordination cost per waiting period. A submitter proceeds if

$$\pi > c_0 + wTk + \kappa. \qquad (7)$$

For a submitter who wants $n$ exceedance submissions, total cost is $n(c_0 + wTk + \kappa)$: linear in $n$. Spreading the waiting across $m$ employees shortens the calendar time required but does not reduce the total cost, because every submission still requires its own period of identity-verified time. Delegation to hired proxies is therefore possible, but it does not change the cost structure: the pathway does not aim to make delegation impossible, only to keep a non-zero, identity-bound marginal cost attached to each exceedance submission, so that the volume passing through it is bounded by the verified time of the people engaged. Deterrence therefore depends on the value of each submission. Strategies whose value lies in volume, in which each item has a low expected payoff but replication is free, such as mass speculative filings or manufactured manuscripts, are deterred once $wTk + \kappa$ exceeds the low per-item payoff. Submissions whose expected value is high are not deterred by any proportionate waiting time. With $\pi = 10{,}000$, $w = 30$ per hour, $k = 2$ and negligible $c_0$ and $\kappa$, for example, the base time would have to exceed 166 hours. For such submissions, the pathway acts as a price rather than a barrier, and well-resourced submitters can pay it, including by employing others to wait.

The design does not depend on deterrence for its central guarantee. Reviewer load is bounded by the cap whatever submitters choose (Proposition 2(i)). Friction contributes three further effects: it removes low-value volume, which under compound growth is the volume that would otherwise fill the queue; it orders admission by willingness to bear verified time rather than by the speed of digital submission, which in a first-come, first-served queue favours mass submitters; and it converts the near-zero marginal cost of digital replication into a labour cost that grows linearly with volume. What friction does not do is equalize access between well-resourced and poorly resourced submitters. Section 7 examines the resulting distributional questions.

### 5.4 Practical difficulties

Identity-bound waiting faces serious practical difficulties. It imposes disproportionate burdens on applicants with disabilities, applicants in remote regions and applicants in low-income countries; accommodation mitigates but does not remove these burdens. Many jurisdictions guarantee prompt intake of applications, and mandatory waiting may require legislative or treaty change. If one jurisdiction introduces the mechanism while another does not, applicants may move between them. For these reasons, the contribution of this paper lies in the four invariants, and any mechanism that satisfies them at lower social cost should replace identity-bound waiting. A plausible route to implementation is for private platforms and journals, which already face floods of AI-generated submissions and already experiment with per-submitter limits [21], to adopt the less burdensome components first, namely identity-bound limits on submissions, the cost score and an explicit capacity cap, so that their effects can be observed before public institutions consider legislative change. Identity-bound waiting is not part of this recommendation.

## 6 Uncertainty Analysis

This section asks under what parameter conditions Flow-by-Flow buys more time than supervision enhancement alone (RQ4). The model is deterministic once its parameters are fixed; the Monte Carlo step propagates uncertainty about the parameters. The analysis does not predict when any particular domain will exceed its capacity, and its results are conditional on the assumed parameter ranges.

## 6.1 Model

Initial capacity is normalized to one, so that load is expressed as a share of initial capacity, and time is measured in years, $t = 0,1, \dots ,30$. Three strategies are compared:

(a) supervision enhancement only: $W_a(t) = u_0(1 + g_V)^t(1 + g_L)^t, \quad C_a(t) = 1 + a_a t, \qquad (8)$

(b) with an identity-bound limit on submissions: $W_b(t) = u_0 \min\{(1 + g_V)^t, m\}(1 + g_L)^t, \quad C_b(t) = 1 + a_b t, \qquad (9)$

(c) with the cost score added: $W_c(t) = W_b(t)\,[1 - s_0 \max(0,1 - t/D)], \quad C_c(t) = 1 + a_c t. \qquad (10)$

Here $u_0 = 1/s$ is initial utilization, $m$ is the factor by which the limit on submissions allows volume to rise above its initial level, $s_0$ is the share of load initially suppressed by the cost score, and $D$ is the number of years over which this suppression decays linearly to zero as applicants learn to optimize against the score. The outcome for each strategy is the first year in which load exceeds capacity, $\tau_k = \min\{t: W_k(t) > C_k(t)\}$, recorded as 31 if no exceedance occurs by year 30. Under strategy (a), $\tau_a$ marks the onset of nominal review. Under (b) and (c), $\tau_k$ marks the year from which the capacity cap must ration admission, because reviewer load itself remains bounded (Proposition 2).

All strategies are evaluated on the same parameter draw. Strategy (c) is said to outlast (a) in a draw if $\tau_c > \tau_a$; ties are reported separately; and the gain is $\Delta = \tau_c - \tau_a$. In the comparison of all three strategies, a strategy is counted as best only if its $\tau$ is strictly larger than those of the other two; ties for best are reported separately.

All parameters are drawn independently from uniform distributions, with no correlations imposed. Strategies (b) and (c) are given slower capacity growth than (a) to represent the possibility that operating flow control diverts resources from supervision enhancement. The range for $g_V$ is an assumption: the increases of 23.7% to 89.3% reported by Kusumegi et al. [18] are one-time increases in the output of adopting authors, not annual growth rates of aggregate volume, and translating them into annual rates requires an assumption about the pace of adoption. If adoption spreads over a few years, aggregate growth of 10% to 27% per year during diffusion is plausible but not established. Growth rates also differ widely across fields. From January to August 2026, mathematics submissions to arXiv were 33.5% above the same period of 2025 [58]. These are preprints rather than submissions to accountable review, but if such output reaches journals, review demand in the fields concerned could grow faster than the 10% to 27% a year assumed above. We therefore also report four domain scenarios for the growth of volume: stable (0% to 5% a year), moderate (the range of Table 5), rapid (50% to 100% a year) and a burst in which volume doubles in each of two years and then grows at the moderate rate. Two structural variants are reported with them. In the first, the score's diversion of capacity growth ends when the score lapses, after which strategy (c) grows capacity at the rate of (b), since an institution would stop paying for a score that no longer works. In the second, the identity-bound limit on submissions diverts no resources from supervision enhancement. There is no direct evidence on $g_L$ (Section 2.2), so its range is wide and the case $g_L = 0$ is analysed separately. The suppression parameters $s_0$ and $D$ have no empirical anchor because the design has not been implemented; for this reason we report a break-even analysis over these two parameters rather than relying on their assumed ranges.

**Table 5** Parameters of the uncertainty analysis (uniform distributions unless stated)

| Parameter | Meaning | Range |
|---|---|---|
| $u_0$ | Initial utilization of capacity | 0.10 to 0.50 |
| $g_V$ | Annual growth of volume | 0.10 to 0.27 |
| $g_L$ | Annual growth of per-item load | 0.10 to 0.40 |
| $a_a$, $a_b$, $a_c$ | Linear growth of capacity under (a), (b), (c) | 0.10 to 0.20; 0.05 to 0.13; 0.02 to 0.08 |
| $s_0$ | Initial suppression of load by the score | 0.50 to 0.95 |
| $D$ | Years until the suppression has decayed | 5 to 50 |
| $m$ | Volume allowed by the limit on submissions, relative to its initial level | 2 (fixed) |

Main results use 100,000 draws with a fixed random seed. To show the Monte Carlo variation at a sample size of 1,000 draws, the analysis was repeated 500 times. The break-even analysis fixes $s_0$ and $D$ on a grid of 20 by 15 values and draws the other parameters 20,000 times at each grid point. Sensitivity is summarized by partial rank correlation coefficients (PRCC) between each parameter and the gain $\Delta$. The code is available as described in the Statements and Declarations.

### 6.2 Results

Table 6 summarizes the results, and Figures 2 to 4 show the main distributions.

**Table 6** Comparison of strategy (c) with strategy (a): share of draws in which (c) outlasts (a), ties, and (a) outlasts (c), with the median gain $\tau_c - \tau_a$ in years and its interquartile range (IQR)

| Scenario | (c) outlasts (a) (%) | Tie (%) | (a) outlasts (c) (%) | Median gain (IQR) |
|---|---|---|---|---|
| Ranges of Table 5 | 91.2 | 7.0 | 1.8 | 3 (1 to 4) |
| Pessimistic: $s_0 = 0.50, D = 5$ | 19.9 | 48.7 | 31.4 | 0 (−1 to 0) |
| Optimistic: $s_0 = 0.95, D = 50$ | 100.0 | 0.0 | 0.0 | 6 (5 to 8) |
| No limit on submissions ($m = \infty$) | 76.0 | 17.5 | 6.5 | 1 (1 to 2) |

| Scenario | (c) outlasts (a) (%) | Tie (%) | (a) outlasts (c) (%) | Median gain (IQR) |
|---|---|---|---|---|
| Limit on submissions only ($s_0 = 0$) | 16.1 | 29.1 | 54.9 | −1 (−1 to 0) |
| Tighter limit ($m = 1.25$) | 99.3 | 0.6 | <0.1 | 4 (3 to 7) |
| Looser limit ($m = 3$) | 81.7 | 13.8 | 4.5 | 2 (1 to 3) |
| No diversion of capacity growth ($a_c = a_a$) | 99.3 | 0.7 | 0.0 | 5 (3 to 7) |
| No growth of per-item load ($g_L = 0$) | 94.5 | 5.5 | 0.0 | ≥16 (≥10 to ≥20)[a] |

[a] With $g_L = 0$, strategies (b) and (c) never exceed capacity within 30 years, so the gain is right-censored and the reported values are lower bounds.

The results support seven findings.

**Finding 1 (baseline).** Under the parameter ranges of Table 5, strategy (c) postpones the first exceedance of capacity relative to (a) in 91.2% of draws, ties in 7.0% and is outlasted in 1.8%. The median gain is three years (IQR 1 to 4). The median exceedance year is 5 under (a) and (b) and 8 under (c) (Figs. 2 and 3). Counting only strict wins among the three strategies, (c) is best in 82.4% of draws, (b) in 6.5% and (a) in 1.1%, with ties for best in 10.0%. With 1,000 draws, the share of draws in which (c) outlasts (a) varies between 89.4% and 93.0% across 95% of replications.

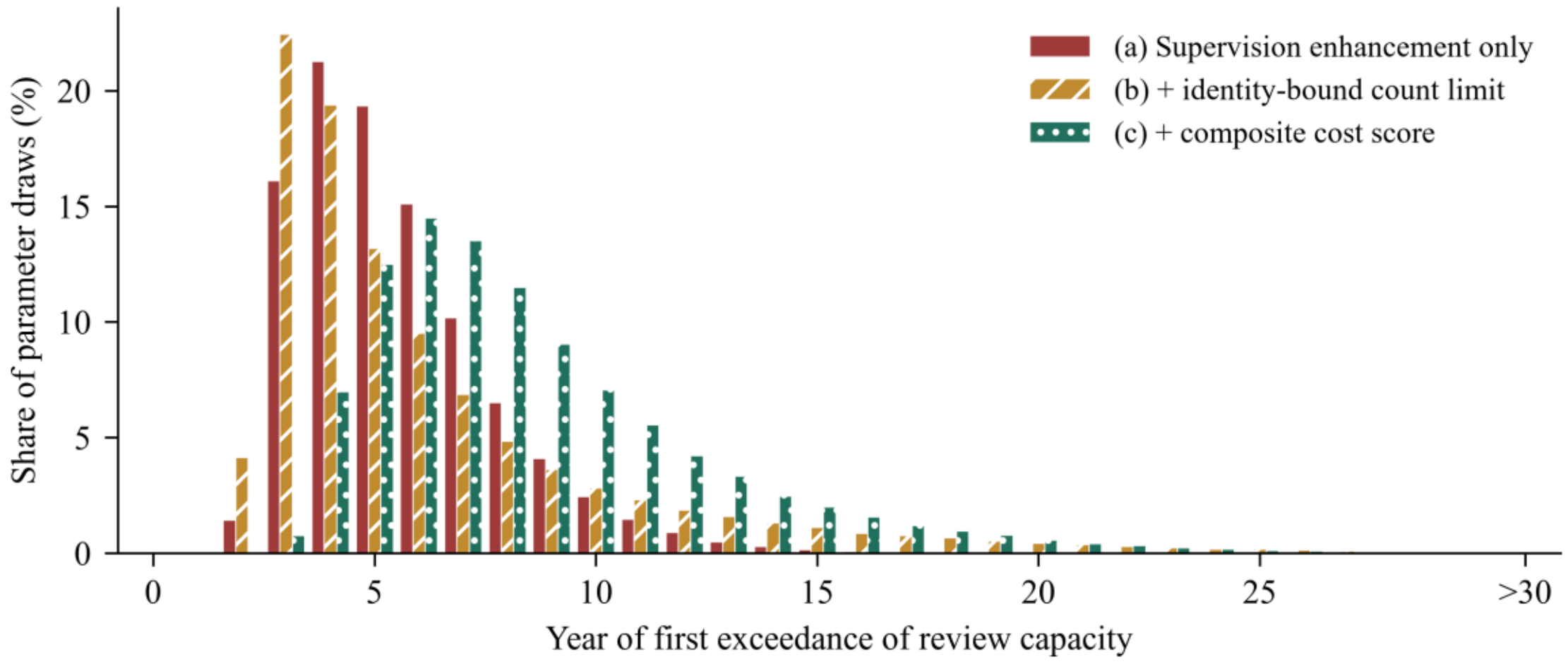

**Fig. 2** Distribution of the first year in which load exceeds capacity under strategies (a), (b) and (c), for 100,000 parameter draws from the ranges of Table 5. The bin labelled ">30" contains draws without exceedance within 30 years; it is almost empty for all strategies

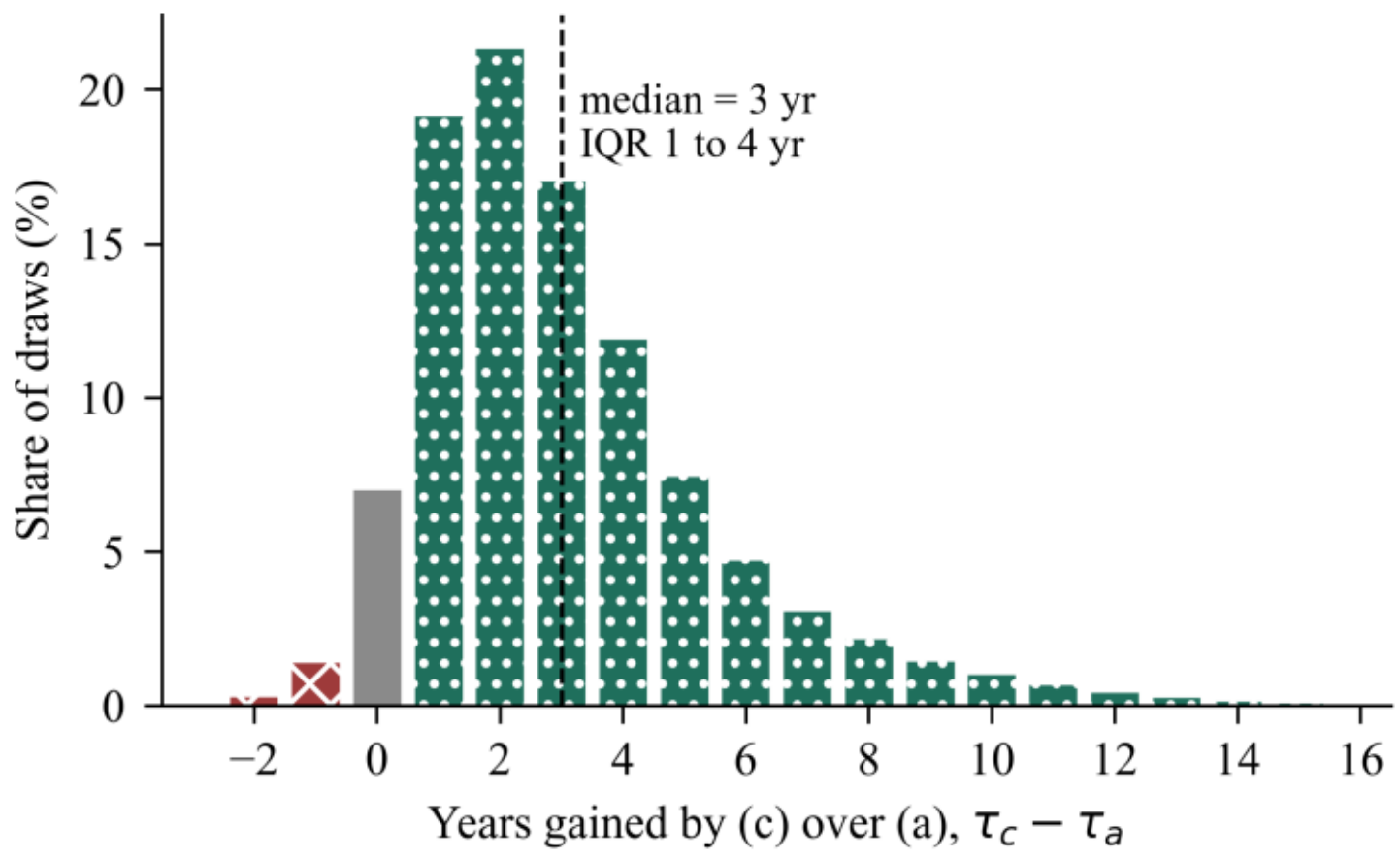

**Fig. 3** Distribution of the gain $\tau_c - \tau_a$, the number of years by which strategy (c) postpones the first exceedance of capacity relative to strategy (a), for the draws of Fig. 2. Dotted bars: (c) outlasts (a); grey bar: tie; hatched bars: (a) outlasts (c). The dashed line marks the median

**Finding 2 (dependence on the suppression parameters).** The advantage of (c) depends on the two parameters that have no empirical anchor. In the pessimistic case of $s_0 = 0.50$ and $D = 5$, (c) outlasts (a) in only 19.9% of draws and is outlasted in 31.4%: the ordering is reversed. The break-even analysis (Fig. 4) shows that if the score's effect decays within about six years, (c) outlasts (a) in fewer than half of the draws whatever its initial strength. With an initial suppression of 50%, the effect must last about ten years for (c) to outlast (a) in a majority of draws, and with an initial suppression of 30%, about forty years.

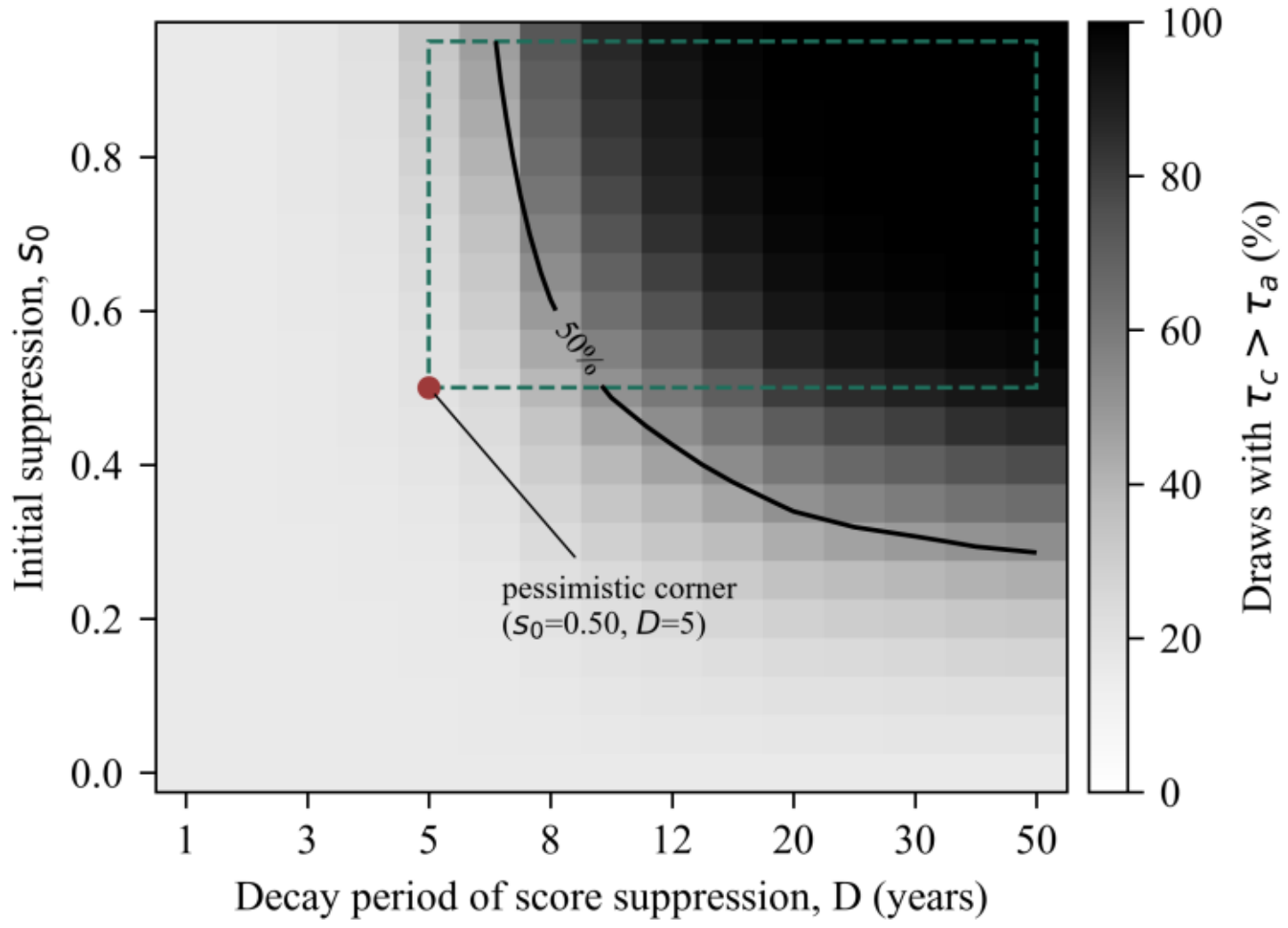

**Fig. 4** Break-even analysis. Shading shows the share of draws in which strategy (c) outlasts strategy (a) as a function of the initial suppression $s_0$ and the decay period $D$, with the other parameters drawn from the ranges of Table 5. The solid line marks 50%. The dashed rectangle marks the ranges assumed in Table 5, and the dot marks the pessimistic case

**Finding 3 (sources of the advantage).** The gain has two sources that act differently, as Corollaries 1 and 2 predict. The limit on submissions acts on the growth rate: once it binds, load grows at $g_L$ rather than $g$. The score acts on the level of load, and its effect fades. Neither suffices alone. With the limit on submissions but no score, the slower capacity growth assumed for (c) makes it worse than (a) in most draws (16.1% versus 54.9%). With the score but no limit on submissions, (c) outlasts (a) in 76.0% of draws with a median gain of one year. With both, the median gain is three years, and a tighter limit raises it to four.

**Finding 4 (sensitivity).** The gain is most sensitive to the decay period $D$ (PRCC 0.74), the growth of per-item load $g_L$ ($-0.71$), the initial suppression $s_0$ (0.59) and the growth of volume $g_V$ (0.52), followed by the capacity growth of (c), $a_c$ (0.36), initial utilization $u_0$ ($-0.30$) and the capacity growth of (a), $a_a$ ($-0.26$). The interaction between the two growth rates is consistent with Finding 3: the share of draws in which (c) outlasts (a) rises from about 80% where volume grows most slowly to above 95% where it grows fastest, and where volume grows fast it declines as per-item load grows faster, because the limit on submissions does not act on $L$ (Appendix C, Table 8). The assumption that flow control diverts resources from supervision enhancement costs (c) about two years: without it, the median gain is five years.

**Finding 5 (censoring).** Under the ranges of Table 5, fewer than 0.04% of draws avoid exceedance within 30 years under any strategy, so the treatment of censored draws does not affect the results. Censoring matters only when per-item load does not grow. In that case strategies (b) and (c) never exceed capacity within 30 years, whereas (a) does so in 94.5% of draws: if only volume grows, a limit on submissions suffices to keep load within capacity.

**Finding 6 (the burden of rationing).** Postponing exceedance does not remove the need to ration. In the median draw, the share of demand that exceeds capacity in year 10 is 82% under (a), 63% under (b) and 50% under (c), and in year 20 it exceeds 90% under all three strategies. Under (a) this excess is reviewed nominally or not at all; under (b) and (c) it must be rationed through the exceedance pathway, quotas or a growing queue (Proposition 2). This burden is the central cost that Section 7 weighs.

**Finding 7 (domains and the robust core).** The advantage of flow limitation depends on how fast volume grows (Appendix C, Table 9). Where volume is stable, the limit on submissions rarely binds, and (c) outlasts (a) in only 48.6% of draws because the score's cost slows the growth of capacity; limits on volume are not needed in such fields. Where volume grows at 50% to 100% a year or in a burst, supervision enhancement alone fails within two years in the median draw, whereas (c) lasts eight years, and even if the score is evaded within two years and its cost ends when it lapses, (c) outlasts (a) in 64.8% and 74.2% of draws respectively. In a field of moderate growth, by contrast, a score evaded within two years does not repay its cost: (c) is then outlasted by (a) in 43.4% of draws and outlasts it in 24.7%. The identity-bound limit on submissions is the robust part of the design. If it diverts no resources from supervision, it never does worse than supervision enhancement alone, and in rapidly growing fields it outlasts it in about 80% of draws, with a median gain of three years.

### 6.3 Interpretation

The analysis supports a qualified conclusion. Under the assumed ranges, combining a limit on submissions with the cost score postpones the onset of overload by a few years relative to supervision enhancement alone. This ordering is robust to Monte Carlo variation but not to the assumptions about the score: it holds only if the score's effect on load lasts for roughly a decade or longer, which cannot be known until the design is tried. The share of draws favouring (c) summarizes the assumed ranges and is not evidence that the design works. Given the speed at which constraint-satisfying rewriting has improved, applicants may learn to optimize against the score within one or two years rather than a decade; in that case the score contributes little (Fig. 4), and the protection of review rests on limits on submissions and the capacity cap.

The analysis also shows where a durable advantage must come from. By Corollary 1, a suppression of load buys time only on a logarithmic scale, about $\ln(1/(1-s_0))/\ln(1+g)$ years if it is permanent and less if it fades; by Corollary 2, controls that change the growth rate of admitted load, such as identity-bound limits on submissions and the capacity cap itself, are what make a lasting difference. The score's role is to keep the cap meaningful by keeping estimated load close to actual load for as long as possible. Finally, the model omits effects that cut in both directions: it ignores the packing of more content into each submission under a limit on submissions, which would weaken (b) and (c), and it ignores the deterioration of detection under overload, which would weaken (a). We report these omissions without adjusting the model for them.

Taken together, the findings suggest an order of adoption. The identity-bound limit on submissions and the capacity cap form the robust core: their value grows with the growth of volume, and they do no harm where volume is stable provided that they divert no resources from supervision. The cost score is an adaptive addition, worth its cost only where limits on submissions are followed by denser submissions and only for as long as it resists evasion; once it no longer does, it should be withdrawn and its resources returned to supervision.

## 7 Ethical Assessment: Costs, Benefits and Rights

Any exceedance pathway that satisfies the invariants imposes friction on access to institutions that people rely on to protect inventions, resolve disputes and evaluate research. Section 5 used identity-bound waiting only to show that the invariants can be met jointly; the questions examined here arise for any mechanism that binds a cost to verified identity and to each submission. This section therefore states them as criteria that any implementation must meet: how its costs compare with its uncertain benefits, proportionality, errors in the estimation of load, privacy, due process and timely access, and unequal burdens.

### 7.1 Weighing costs against benefits

The most serious objection to the design is that its costs fall partly on disadvantaged applicants, while its benefits, a few more years of substantive review, depend on parameters that cannot yet be estimated. Three considerations bear on this comparison. None of them settles it in general.

First, the costs of the design are contingent on the same conditions as its benefits. The exceedance pathway is used only by submissions beyond ordinary bounds, and the cap rations admission only when demand exceeds capacity. If supervisory load does not grow, the pathway is rarely used and the cost of the design is essentially the cost of operating the automatic measurement layer. The friction grows with the overload that makes protection valuable (Finding 6). Both the cost and the benefit therefore scale with the unknown growth of load.

Second, the alternative to explicit rationing is not unimpeded access. Under persistent overload, review becomes cursory: submissions are processed in batches with little examination, or judged by signals other than their content, such as affiliation or reputation (Appendix A.3). Submitters who lack favourable signals then bear a further cost, and undetected errors impose costs on third parties, such as patients, litigants and users of the scientific record, who never appear before the institution. These costs also fall disproportionately on the disadvantaged, although, unlike the costs of the exceedance pathway, they are diffuse and unrecorded.

Third, the comparison can be made explicit for a given domain. Adoption is justified only if the expected harm avoided, that is, the yearly harm of nominal rather than substantive review multiplied by the number of years of substantive review gained, exceeds the aggregate cost of friction, including the time of applicants on the exceedance pathway and the additional burden of those who need accommodation, together with the operating cost of the design. Section 6 indicates that the years gained may be few under the assumed ranges and close to zero if the score's effect decays quickly. We do not claim that this condition holds in any particular domain, because the harm and the costs have not been estimated. The condition does, however, specify what an institution would need to estimate before adoption, and it implies an order of measures. Identity-bound limits on submissions, which act on the growth rate, are cheaper and more effective than the score (Finding 3) and suffice when only volume grows (Finding 5). The score and the exceedance pathway are justified only where limits on submissions are followed by increasing load per submission, and the score only for as long as it resists evasion (Finding 7). Among mechanisms that satisfy the invariants, the least burdensome should be chosen.

### 7.2 Proportionality

A measure that burdens access to an institution should pursue a legitimate aim, be suitable for achieving it, be necessary in the sense that no less burdensome measure would achieve it, and impose burdens that are not excessive relative to its benefits. The aim, preserving substantive review in high-loss domains, is legitimate, and the cap is suitable for it because it bounds reviewer load by construction (Proposition 2). Necessity holds only relative to the threat model: monetary friction fails Invariant 3, and content-based screening fails Invariant 1, but where limits on submissions suffice, the exceedance pathway is not necessary.

Proportionality in the strict sense concerns the size of the friction. If friction is proportional to the multiplier $k$ by which a submission's estimated load exceeds the threshold, as in (6), its cost grows with the capacity consumed, which is proportionate in the sense of congestion pricing. It may still be excessive for an individual applicant who legitimately needs a complex submission: with waiting, a submission with $k = 8$ requires sixteen hours. Any implementation should therefore set a maximum friction per submission, allow it to be divided, and offer hardship provisions (Section 7.5). Where the friction needed to keep admitted load within capacity would exceed these limits, the institution should use quotas or lotteries rather than longer waiting (Section 3.5).

### 7.3 Errors in the estimation of load

Because the design does not classify intent, a "false positive" in the usual sense does not arise: a legitimate submission that genuinely imposes a high load is priced correctly, not misclassified. The relevant error is a score that overstates the actual load of a submission. Such errors can be measured, because review time is recorded after each review. Two remedies follow. If the score systematically overstates load for a field or a feature, the weights are recalibrated (Section 3.3). If an individual submission required much less review time than its score implied, the submitter can receive a credit, for example against future exceedance submissions. Statistics on these errors should be published, so that the score can be contested on evidence rather than on its formula alone.

### 7.4 Privacy and data protection

Binding friction to verified identity creates personal data, whatever the mechanism. Identity checks should therefore follow data minimization and purpose limitation [59]: the record retained should be limited to a pseudonymous link between a verification and a submission, kept only as long as audits require and not shared across institutions (Appendix B.1). Where verified identity is not already required for submission, as in much of academic peer review, introducing it requires separate justification.

### 7.5 Due process and timely access

The design distinguishes formal access to an institution, meaning the right to submit and the legal date of a submission, from timely access to a decision. The legal date is secured at filing, so that friction affects the order and timing of processing but not rights that depend on dates, such as priority in patent law or compliance with limitation periods. Some proceedings are time-sensitive in themselves, for example statutory decision deadlines, urgent interim relief or time-critical approvals. For these, delay can affect substantive outcomes, and they require exemption or priority lanes defined by the formal type of proceeding, not by content.

Timely access to a decision is also protected by law. Intellectual property procedures must allow rights to be granted within a reasonable period and must not involve unwarranted delays [60], and the rights to a fair hearing within a reasonable time and to good administration apply to courts and, in the European Union, to administrative bodies [61, 62]. These obligations are the reason why the bounded backlog of Proposition 2 matters: an unbounded queue would breach them, whereas an explicit cap with regulated admission allows an institution to state, and be held to, the time within which admitted submissions are decided. Where existing law guarantees immediate processing, the design would require legislative change, which should be debated as such.

Domains also differ in which controls are legitimate. A per-submitter cap of the kind that a journal or a patent office can impose would be far harder to justify for parties to litigation, for whom access to a court is a right rather than a service. Courts already govern the volume and length of filings by other means: appellate briefs are limited in length [63], a person found to be a vexatious litigant may file new litigation without a lawyer only with the leave of a presiding judge [64], and lawyers certify by their signature that the legal and factual contentions of a filing are supported [65]. The second of these relies on a judge's assessment of merit and so does not meet Invariants 1 and 2, which shows how scarce judicial time already shapes access to courts. In courts, the capacity that most needs protection is that of judges: a judge who, under load, delegated the drafting of decisions to AI without scrutiny would introduce errors at the point of decision itself.

Applicants also need recourse. Routing decisions rest on counts, so errors in routing can be contested mechanically by recounting. Hardship waivers for applicants who cannot bear the friction, because of disability, distance or income, can be decided by staff who do not review content, using formal criteria. To remain compatible with Invariants 3 and 4, waivers should themselves be identity-bound and limited per period, so that they cannot become a route to clearance in bulk.

### 7.6 Unequal burdens

Identity-bound friction does not burden everyone equally. With waiting, for example, applicants with disabilities, applicants who live far from an office and applicants in low-income countries face higher costs, which accommodation (Section 5.2) reduces but does not remove. Friction in time is distributed more evenly than friction in money, because everyone has the same number of hours. The opportunity cost of time is not equal, however, and the literature on targeting and administrative burden shows that burdens of this kind tend to fall most heavily on people with the fewest resources [37, 66]. Friction in time is therefore a burden that requires justification. Conversely, well-resourced submitters can hire others to wait (Section 5.3). Under the logic of congestion pricing this is acceptable in principle, because they pay for the capacity they consume, but it means that the design does not equalize access between rich and poor submitters. Per-submitter allowances, which apply equally to everyone, and accommodation for those who cannot wait are therefore part of the design itself.

Finally, the design must be administratively feasible. Identity verification has a limited capacity (Section 3.5), offices and staff have costs, and coordination across jurisdictions is needed to prevent applicants from moving between institutions. These requirements are one reason why private platforms and journals, which can introduce limits on submissions and capacity caps without legislation, are the more realistic first adopters (Section 5.4).

## 8 Limitations

### 8.1 Status of the framework and its calibration

The contribution of this paper is theoretical. The inequality $V \times L \leq C$ is an uncalibrated hypothesis about institutional capacity, and the cognitive cost score is a proposed proxy for load. Individual features such as the number of claims are known to lengthen examination (Section 3.3), but the relationship between the composite score and actual review effort has not been measured. Neither should be treated as an established measure. Each could be calibrated from records that many institutions already keep. Review capacity can be estimated from the relationship between the volume handled per reviewer and error rates, such as missed defects, incorrect approvals and later reversals, with the cap set below the point at which error rates begin to rise; periodic audits that insert known errors can detect later deterioration of detection. The weights of the score can be estimated by regressing recorded review time on the normalized features, and the floored product (5) can be tested against the additive alternative by out-of-sample prediction. The growth of per-item load can be tracked by monitoring review time per submission over time, and the staggered introduction of controls across units or fields would permit before-and-after comparisons. Until such calibration has been done, the framework identifies what to measure and why, but not the values that an institution should adopt.

### 8.2 Features of the score

The features used as examples, such as the numbers of claims, words and references, are initial approximations. Their resistance to evasion differs: the number of references, for example, may be easier to reduce strategically than the length of a text. Adding features makes evasion harder but increases the chance that legitimate submissions in specialized fields exceed the threshold. The number of features should therefore be chosen by weighing the difficulty of evasion against errors in the estimation of load, and revised as operational data accumulate.

### 8.3 Growth of per-item load

The evidence for the growth of volume is stronger than the evidence for the growth of per-item load. The mechanisms through which per-item load may grow, including the cost of triage and uncertainty about provenance (Section 2.2), are hypotheses, and the range assumed for $g_L$ is not estimated from data. For this reason Section 6 reports results for the case in which per-item load does not grow. The argument as a whole requires that supervisory load grow faster than capacity. It would lose its force if the use of generative AI contracted, if its capabilities stopped improving, or if a growing share of outputs in the domains concerned could be verified mechanically, so that errors in that share could be excluded.

### 8.4 The simulation

The model of Section 6 is deliberately simple. Its parameters are drawn independently from uniform distributions; the suppression parameters $s_0$ and $D$ have no empirical anchor; time is discrete; and the model omits several effects, including the packing of more content into each submission under a limit on submissions, the deterioration of detection under overload, the dynamics of the queue, and the strategic responses of institutions. Its results should be read as statements about the consequences of the assumed ranges, not as estimates.

### 8.5 Jurisdictions

The analysis takes a single institution as its unit. Patent offices, drug regulators and publishers are linked by mutual recognition and competition, and if one adopts flow control while others maintain full immediate intake, applicants may move between them. Coordination among institutions is outside the scope of this paper.

### 8.6 The endpoint of evasion

Flow-by-Flow postpones but does not remove the point at which countable features cease to distinguish submissions. As optimization capability grows, every countable dimension can eventually be saturated, and the remaining differences between submissions lie in qualities that only content judgment can assess. At that point the problem that the design bypasses returns. Its value therefore depends on what institutions do with the time it buys, for example developing mechanical verification where it is feasible, reforming the allocation of accountability, or redesigning the processes that generate the load in the first place.

## 9 Conclusion

Human review remains the layer at which AI outputs are accepted and to which accountability is anchored in high-loss domains. This paper has asked how that layer can be protected when the volume and complexity of submissions grow faster than review capacity, without asking reviewers or AI systems to judge content.

The answers to the four research questions can be summarized as follows. First, if supervisory load grows at a compound rate while capacity grows linearly, supervision enhancement alone fails in finite time; capacity investment and one-off reductions of load postpone this only logarithmically, whereas controls that lower the growth rate of admitted load postpone it hyperbolically (Proposition 1). Second, a mechanism that admits submissions beyond ordinary bounds must, under the stated threat model, avoid content judgment and the scalable use of reviewer time, and must attach an identity-bound cost to each submission without permitting clearance in bulk. Third, these requirements can be satisfied jointly, as identity-bound waiting shows, but at costs in proportionality, privacy, due process and the distribution of burdens that require safeguards and that justify adoption only where the expected harm of nominal review is large. Fourth, under the assumed parameter ranges, combining a limit on submissions with the cost score postpones the onset of overload by a median of three years relative to supervision enhancement alone, and this advantage disappears if the score's effect decays within about six years. The durable part of the advantage comes from controls that act on the growth rate of admitted load.

These results do not diminish the value of strengthening human oversight, hiring reviewers or improving their tools. They show that such measures act on the level of capacity and therefore cannot keep pace with compound growth on their own. Flow-by-Flow complements them by bounding the load that reaches accountable reviewers and by converting overload from an invisible collapse of review quality into an explicit allocation that can be governed and contested. It is a mitigation that buys time, not a permanent remedy. Its most realistic first adopters are private platforms and journals, which already face floods of AI-generated submissions and can test limits on submissions, the cost score and capacity caps without legislative change; public institutions can then consider the legal reforms that adoption would require on the basis of evidence rather than projection.

#### Statements and Declarations

**Competing interests.** The author has no competing interests to declare that are relevant to the content of this article.

**Funding.** No funding was received for conducting this study.

**Ethics approval and consent.** Not applicable.

**Data availability.** No empirical data were collected for this study.

**Code availability.** The simulation code (Python, with a fixed random seed) and its output files have been provided to the editors in anonymized form as supplementary material and will be made publicly available upon acceptance.

**Use of AI tools.** AI tools were used for translation, development of simulation code, and formatting and visual refinement of figures and tables.

## Appendix A: Responses to Standard Objections

### A.1 Accuracy improvements and the need for flow limitation

A common objection is that supervisory load will fall as models become more accurate. Three considerations weigh against it. First, the error probability of a probabilistic generator cannot be driven to zero in the domains considered here [67], and a system that answers consistently across equivalent formulations of a problem cannot avoid errors unless it can abstain [68]. Second, only the judgment component of per-item load is under downward pressure from accuracy (Section 2.2). Response work scales with the number of accepted outputs and is independent of accuracy. If response work is delegated to AI, the result of that delegation must itself be verified, because accountability rests with a person ($R = 1$); delegating the verification starts a recursion that ends with a human reviewer, to whom the capacity constraint applies. Third, accuracy improvements make the remaining errors rarer, and rare targets are disproportionately missed by human inspectors [31], so that the cost of verifying highly accurate output can rise rather than fall.

If AI-assisted development triples the number of applications to a review agency while better screening halves its error rate, harmful approvals rise by half: each review is more accurate, yet the system is more dangerous. Bastani and Cachon [69] show, in a complementary analysis, that the compensation needed to keep supervisors vigilant rises as AI accuracy improves; the present analysis adds that even motivated supervisors face a capacity constraint.

## A.2 AI-use disclosure regimes

Requirements to disclose whether AI was used cannot be enforced unless the disclosure can be verified, and no reliable method exists for distinguishing AI-generated from human-written text after the fact. Widely used detectors misclassified more than half of essays by non-native English writers as AI-generated, while simple prompting reduced their detection rates sharply [70], and a text classifier released by a major AI developer was withdrawn because of its low accuracy [71]. Where disclosure cannot be verified, reviewers must assume that any submission may be AI-generated, so that the spread of AI raises review costs even for submissions produced without it.

## A.3 Consequences of nominal oversight

When supervisory load exceeds capacity, institutions do not stop. Reviewers continue to approve and report throughput as before, and from the outside the institution appears to function, but what continues is a flow of formal approval rather than substantive review. Review can then degrade in more than one way. Some submissions are approved or rejected in batches with little or no examination; others are judged by signals other than their content, such as the applicant's affiliation, reputation or representation. In either case the costs of nominal review fall on third parties affected by undetected errors, and the second pattern adds a further cost for submitters who lack favourable signals.

## A.4 Limits of content judgment

In current debates, human oversight almost always means content judgment: deciding whether a text is accurate, code is secure or a document is legally sound. Recent cases show that content judgment fails even when it is performed by qualified people. Since a lawyer filed a brief citing cases invented by a chatbot in Mata v. Avianca [72], courts have issued a growing number of decisions addressing fabricated authorities in filings [73]. In one appellate case, non-existent cases cited in a filing passed both the filing lawyer and the trial court, whose order relied on them before it was vacated on appeal [74]. In peer review, fabricated references passed the reviewers of papers accepted at leading conferences [23]. In clinical documentation, summaries generated by language models contained hallucinations in 1.47% and omissions in 3.45% of cases [75], rates low enough to escape attention item by item but high enough to accumulate in high-volume use. Errors need not be fabrications: a summary consisting only of statements found in its source can still reverse the source's meaning through what it omits.

Delegating content judgment to AI does not remove these limits. Detectors of AI-generated text are unreliable (Appendix A.2), retrieval augmentation, cross-checking by several models and prompt engineering reduce errors without eliminating them, and the correctness of an AI judgment must in turn be established. For this reason the design in this paper does not seek better content judgment at the point of admission, but avoids content judgment there altogether.

## Appendix B: Further Discussion of the Exceedance Pathway

### B.1 Other forms of friction

Several other mechanisms could limit the volume of submissions, but each fails at least one invariant of Section 3.6. Deposits and other forms of economic staking are monetary friction, which fails Invariant 3 (Appendix B.2), fails Invariant 4 as well when a single deposit covers several submissions, and discriminates against applicants with few resources. Restricting submission to qualified persons does not limit the volume produced by qualified persons who use AI. Penalties imposed after the fact by a single institution can be evaded by moving to another institution, and preventing this would require institutions to share identity and penalty records, a surveillance infrastructure that Section 7.4 rejects. These observations do not establish that identity-bound waiting is preferable to all alternatives; they indicate the conditions under which each alternative fails.

### B.2 Monetary friction as a precursor

Fees on formal features already exist (Section 3.3). After the European Patent Office introduced excess-claims fees in 2008, claim counts fell while the breadth of protection was maintained through reduced claim depth [41]. Under a fee schedule that prices claims beyond twenty but leaves specification length effectively unpriced, specifications have lengthened steadily for more than two decades [42], and practitioner guidance recommends combining separate claims into alternative embodiments within a single claim to avoid fees [43]. Monetary friction nevertheless fails Invariant 3. A fee is paid in money, which any account can supply and which a well-capitalized submitter can spend on any number of submissions without engaging anyone else. Because wealth differs across submitters by orders of magnitude while the hours available to each person do not, a fee high enough to limit a well-capitalized submitter would exclude many others. Identity-bound time can also be obtained by engaging others to wait, but each of them spends their own verified time, so the volume passing through the pathway grows only with the number of people mobilized. A fee charged per submission keeps a positive cost for each additional submission and so satisfies Invariant 4; a flat fee, a subscription or a volume discount, under which one payment covers additional submissions at little or no extra cost, does not. For a high-volume submitter whose expected return per submission exceeds the fee, a fee acts as a tax on volume rather than a limit. Fees can be a component of flow design, but not a substitute for an identity-bound exceedance pathway.

## Appendix C: Further Results of the Uncertainty Analysis

Table 7 compares all three strategies in the main scenarios, Table 8 reports the interaction between the growth rates of volume and per-item load, and Table 9 reports the domain scenarios and structural variants described in Section 6.1. All results are produced by the code described in the Statements and Declarations with a fixed random seed.

**Table 7** Comparison of the three strategies: median year of first exceedance, share of draws in which each strategy alone lasts longest, and share of ties for best

| Scenario | Median $\tau_a$ | Median $\tau_b$ | Median $\tau_c$ | (a) best (%) | (b) best (%) | (c) best (%) | Tie for best (%) |
|---|---|---|---|---|---|---|---|
| Ranges of Table 5 | 5 | 5 | 8 | 1.1 | 6.5 | 82.4 | 10.0 |
| Pessimistic: $s_0 = 0.50$, $D = 5$ | 5 | 5 | 5 | 19.2 | 25.0 | 4.5 | 51.3 |
| Optimistic: $s_0 = 0.95$, $D = 50$ | 5 | 5 | 11 | 0.0 | 0.2 | 99.5 | 0.2 |

**Table 8** Share of draws (%) in which strategy (c) outlasts strategy (a), by growth rate of volume $g_V$ (columns) and of per-item load $g_L$ (rows)

| $g_L$ (rows), $g_V$ (columns) | 0.10 to 0.13 | 0.13 to 0.17 | 0.17 to 0.20 | 0.20 to 0.24 | 0.24 to 0.27 |
|---|---|---|---|---|---|
| 0.10 to 0.15 | 83.4 | 92.2 | 96.3 | 98.1 | 99.2 |
| 0.15 to 0.20 | 79.9 | 88.7 | 94.1 | 97.0 | 98.9 |
| 0.20 to 0.25 | 79.5 | 87.0 | 92.8 | 96.6 | 98.2 |
| 0.25 to 0.30 | 80.4 | 88.0 | 91.7 | 95.4 | 97.2 |
| 0.30 to 0.35 | 81.6 | 87.0 | 90.5 | 94.4 | 95.7 |
| 0.35 to 0.40 | 84.6 | 87.8 | 90.8 | 93.5 | 95.1 |

**Table 9** Domain scenarios for the growth of volume. Columns 2 to 4: strategy (c) against (a) under the ranges of Table 5. Columns 5 and 6: (c) against (a) when the score is evaded within two years ($s_0 = 0.50$, $D = 2$) and its diversion of capacity growth ends when it lapses. Columns 7 and 8: the identity-bound limit on submissions (b) against (a) when the limit diverts no resources from supervision

| Scenario | (c) outlasts (a) (%) | (a) outlasts (c) (%) | Median gain (years) | Fast evasion: (c) outlasts (a) (%) | Fast evasion: (a) outlasts (c) (%) | (b) outlasts (a) (%) | (a) outlasts (b) (%) |
|---|---|---|---|---|---|---|---|
| Stable (0% to 5%) | 48.6 | 33.4 | 0 | 0.5 | 88.0 | 2.1 | 0.0 |
| Moderate (10% to 27%) | 91.2 | 1.8 | 3 | 24.7 | 43.4 | 49.4 | 0.0 |
| Rapid (50% to 100%) | 100.0 | 0.0 | 5 | 64.8 | 0.3 | 79.5 | 0.0 |

| Scenario | (c) outlasts (a) (%) | (a) outlasts (c) (%) | Median gain (years) | Fast evasion: (c) outlasts (a) (%) | Fast evasion: (a) outlasts (c) (%) | (b) outlasts (a) (%) | (a) outlasts (b) (%) |
|---|---|---|---|---|---|---|---|
| Burst (100% for two years, then 10% to 27%) | 100.0 | 0.0 | 5 | 74.2 | 0.0 | 79.6 | 0.0 |